\documentclass[letterpaper, 10 pt, conference]{ieeeconf}
\IEEEoverridecommandlockouts
\usepackage{graphicx}
\usepackage{subcaption}
\usepackage{amsmath,amssymb}
\usepackage{hyperref}
\usepackage{xcolor}
\usepackage{comment}
\graphicspath{{figures/}}

\title{\LARGE\bf
Real-Time Decoding of Movement Onset and Offset for Brain-Controlled Rehabilitation Exoskeleton}

\author{Kanishka Mitra$^{1,2}$, Satyam Kumar$^{2}$, Frigyes Samuel Racz$^{3}$, \\ Deland Liu$^{2}$, Ashish D. Deshpande$^{4,5}$, Jos\'e del R. Mill\'an$^{2,3,6}$%
\thanks{$^{1}$Department of Electrical Engineering and Computer Science, Massachusetts Institute of Technology, Cambridge, MA, USA. Email: \texttt{mitra819@mit.edu}}%
\thanks{$^{2}$Chandra Department of Electrical and Computer Engineering, The University of Texas at Austin, Austin, TX, USA. Email: \texttt{\{satyam.kumar, deland.liu\}@utexas.edu, jose.millan@austin.utexas.edu}}%
\thanks{$^{3}$Department of Neurology, The University of Texas at Austin, Austin, TX, USA. Email: \texttt{fsr324@austin.utexas.edu}}%
\thanks{$^{4}$Walker Department of Mechanical Engineering, The University of Texas at Austin, Austin, TX, USA. Email: \texttt{ashish@austin.utexas.edu}}%
\thanks{$^{5}$Meta Reality Labs Research, Redmond, WA, USA.}
\thanks{$^{6}$Department of Biomedical Engineering, The University of Texas at Austin, Austin, TX, USA.}}

\begin{document}

\maketitle
\thispagestyle{empty}
\pagestyle{empty}


\begin{abstract}

Robot-assisted therapy can deliver high-dose, task-specific training after neurologic injury, but most systems act primarily at the limb level---engaging the impaired neural circuits only indirectly---which remains a key barrier to truly contingent, neuroplasticity-targeted rehabilitation. We address this gap by implementing online, dual-state motor imagery control of an upper-limb exoskeleton, enabling goal-directed reaches to be both initiated and terminated directly from noninvasive EEG. Eight participants used EEG to initiate assistance and then volitionally halt the robot mid-trajectory. Across two online sessions, group-mean hit rates were 61.5\% for onset and 64.5\% for offset, demonstrating reliable start–stop command delivery despite instrumental noise and passive arm motion. Methodologically, we reveal a systematic, class-driven bias induced by common task-based recentering using an asymmetric margin diagnostic, and we introduce a class-agnostic fixation-based recentering method that tracks drift without sampling command classes while preserving class geometry. This substantially improves threshold-free separability (AUC gains: onset +56\%, $p=0.0117$; offset +34\%, $p=0.0251$) and reduces bias within and across days. Together, these results help bridge offline decoding and practical, intention-driven start–stop control of a rehabilitation exoskeleton, enabling precisely timed, contingent assistance aligned with neuroplasticity goals while supporting future clinical translation. Project page, code, and supplementary videos: \url{https://mitrakanishka.github.io/projects/startstop-bci/}.

\end{abstract}

\section{Introduction}

Despite decades of effort, robotic exoskeletons have yet to deliver clinically meaningful recovery after stroke. While they provide scalable, high-dose, task-specific therapy, their effects remain modest \cite{mehrholz2018electromechanical}\cite{klamroth2014three}. A central limitation is that most robotic interventions act at the distal, mechanical level, with only indirect influence on the neural circuits that must adapt during recovery. In contrast, therapies gated by neural intent can align efferent drive with afferent feedback, a contingency known to promote activity-dependent plasticity \cite{biasiucci2018brain}. Surface EMG triggers partially embody this idea by coupling assistance to residual muscle activity, but they underperform when paresis is severe \cite{ho2011emg}.

To overcome these limitations, brain-computer interfaces (BCIs) decode motor intention directly from neural activity.
Motor imagery (MI) offers a noninvasive BCI route to intention-contingent control: 
kinesthetically imagining an action—without executing it—elicits event-related desynchronization (ERD) 
over sensorimotor cortex in the $\mu$ (8--13~Hz) and $\beta$ (13--30~Hz) bands measured with scalp 
electroencephalography (EEG) \cite{pfurtscheller2001motor}\cite{tonin2021noninvasive}.
MI-BCIs have been integrated with rehabilitation robots, yet most systems still rely on (i) a single discrete mental command that triggers complex gestures \cite{barsotti2015full}, which restricts the action repertoire, or (ii) maintaining the MI state during robot movement \cite{frolov2017post}, which becomes brittle as proprioceptive feedback and device artifacts mask the MI rhythms being decoded.

A principled approach to intention-contingent robotic assistance leverages the distinct neural signatures of MI onset (Start) and MI termination (Stop): one decoder initiates assistance against Rest, and a second decoder terminates assistance against Maintain-MI \cite
{orset2021stopping}. Detecting offset---not just onset---matters for rehabilitation. First, intention-gated cessation of assistance near a target supports assist-as-needed control, improving precision and safety by avoiding run-on torques and enforcing minimal necessary aid \cite{baud2021review}. Second, reliable stopping supports agency and reduces user confusion---factors linked to engagement and motor learning in technology-mediated movement training \cite{nataraj2020agency}. Third, controlling offset trains inhibitory control, a core motor function associated with better dexterity recovery post-stroke \cite{plantin2022motor}. Finally, unchecked carry-through can reinforce maladaptive agonist--antagonist co-contraction, which is associated with poorer upper-limb function after stroke \cite{chalard2019spastic}\cite{sheng2022upper}. Prior work shows that MI onset and offset are separable and detectable in offline or pseudo-online analyses---including under passive movement---supporting the feasibility of start/stop-style control \cite{orset2021stopping}\cite{mitra2023characterizing}. What has been missing is a real-time demonstration that both transitions can be decoded online to directly start and stop a rehabilitation exoskeleton,
while (i) maintaining accuracy despite movement-evoked sensory feedback and robot operation distorting the signals, and (ii) mitigating EEG drifts that degrade online BCI performance.

\begin{figure*}[ht!]
    \centering
    \includegraphics[width = 7.16 in]{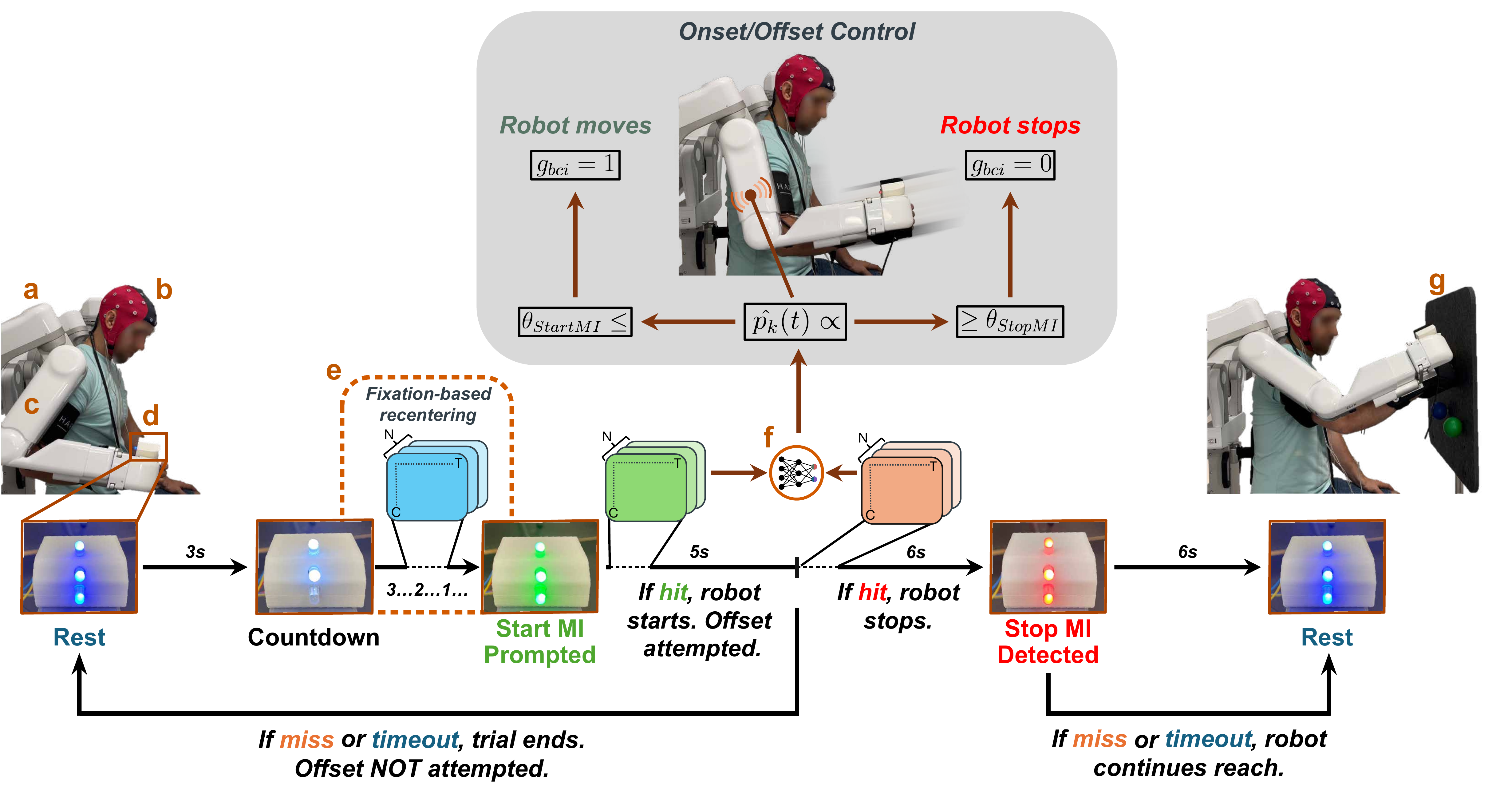}
    \caption{Experimental setup, task timeline, and decoding pipeline for onset/offset EEG control of the Harmony exoskeleton. (a) Harmony (\S\ref{harmony}); (b) EEG cap; (c) stNMES pads under the cuff; (d) arm-mounted LEDs indicating trial phase (white: countdown; green: Start; red: Stop; blue: Rest); (e) fixation-based recentering (\S\ref{pseudo}); (f) subject-specific Riemannian decoders (\S\ref{training}); (g) three spatial targets defining goal-directed trajectories. Blue, green, and orange sliding windows depict EEG samples of shape $N\times C\times T$ (trials $\times$ channels $\times$ time) with T=512 points ($1$ s at $512$ Hz), as used in \S\ref{training}. Onset/offset decisions and robot transitions follow the thresholding logic in \S\ref{threshold} and state machine in \S\ref{state_robot}.}
    \label{fig:setup}
    \vspace{-3mm}
\end{figure*}

We demonstrate online, dual-state MI control to initiate and terminate robotic assistance on an upper-limb exoskeleton, providing a real-time solution to a key barrier in noninvasive BCI-driven therapy. This is implemented using a geometry-aware (Riemannian) EEG pipeline paired with a class-agnostic, fixation-based recentering strategy that tracks EEG drift and significantly improves pseudo-online performance. Our contributions are threefold. (1) To our knowledge, the first online demonstration of dual-state MI control that both initiates and terminates robot assistance during movement using EEG in naïve healthy users. (2) An online Riemannian decoder with subject-specific thresholds and state gating that maintains separability despite exoskeleton-induced variability. (3) Identification and quantification of bias during BCI operation and a technique to reduce it and improve performance. Together, these elements provide a practical foundation for intention-contingent, start/stop assistance aligned with assist-as-needed principles and neuroplasticity goals---positioning MI-BCI with robotic control for successful clinical translation.


\section{Methods}

This section describes the experimental design for EEG-based control of the Harmony exoskeleton, including the participant cohort and task protocols, the apparatus and control architecture, the Riemannian decoding pipeline, and the neurophysiological and behavioral metrics used to evaluate performance.

\subsection{Participants and Protocol}

Eight healthy, right-handed adults (mean age $24.25 \pm 3.01$ years; 1 female) participated after providing written informed consent (IRB protocol 2020-03-0073). This sample size was chosen in line with prior MI-BCI exoskeleton studies, where cohorts of 6-10 participants were sufficient to establish proof-of-concept for online control \cite{mitra2023characterizing,kumar2019towards}. EEG was recorded from 60 scalp locations using a 64-channel eego system (ANT Neuro, Germany) arranged according to the international 10–10 system while participants were strapped into the Harmony upper-limb rehabilitation exoskeleton (\S\ref{harmony}) (Fig. \ref{fig:setup}, a,b). Four channels (T7, T8, M1, M2) were excluded due to artifact susceptibility. Bipolar electrooculography (EOG) was recorded to reject trials with potential ocular contamination as described in \cite{Perdikis2018Cybathlon}. The setup also included three LEDs mounted to Harmony’s right arm (Fig. \ref{fig:setup}, d), whose color indicated the current trial phase. Participants were instructed to maintain visual fixation near the right-arm cue LEDs while attending peripherally to the limb and workspace, supporting somatosensory congruence during MI-driven control \cite{biasiucci2018brain,mitra2023characterizing}. The study comprised three sessions: one offline calibration session with six runs of 20 trials, followed by two closed-loop online sessions described below.

During the offline protocol (Session 1), each trial followed a standardized, LED-cued timeline designed to elicit clear contrasts for Start MI and Stop MI. From a start posture (sitting, right arm by the side, elbow at $\sim90^\circ$, strapped to the exoskeleton), a trial began with 3\,s rest, followed by a 3\,s countdown indicated by sequential white LEDs (1\,s each). Then, the LEDs turned green to cue Start MI; participants intentionally initiated right-arm MI. After 3\,s of Start MI, the exoskeleton began a goal-directed reach toward one of three spatial targets positioned in front of the participant (Fig.~\ref{fig:setup}, g). Targets varied across trials to span distinct movement directions, ensuring that Start/Stop MI decoding did not depend on any single kinematic pattern and demonstrating that the neural signatures generalize across trajectories. For training the Stop MI decoder, ``Maintain'' samples were taken during this movement period (i.e., participants continued Start MI while the robot moved). Three seconds after movement onset, the LEDs switched to red to cue Stop MI for 3\,s; participants actively imagined bringing the ongoing reach to a halt while the exoskeleton continued to move the arm toward the target. The robot then stopped, the LEDs turned blue to indicate rest for 3\,s (participants were instructed not to perform any specific mental strategy), and finally, the exoskeleton returned to the home position, concluding the trial (total duration $\sim$21\,s). This design cleanly contrasts MI initiation with the limb stationary versus MI termination with the limb moving, capturing both MI-related dynamics and movement-evoked reafference under natural human--robot interaction.

To strengthen kinesthetic MI and provide effector-level feedback congruent with the intended action, we delivered sensory-threshold neuromuscular electrical stimulation (stNMES; RehaMove3, Hasomed GmbH, Germany) via surface electrodes on upper-arm muscles (Fig.~\ref{fig:setup}, c). During Start MI, stimulation targeted the triceps brachii (primary elbow extensor engaged in reaching); during Stop MI, stimulation targeted the biceps brachii (antagonist elbow flexor) \cite{Forro2023anatomy}. For each participant, the minimum intensity was set to the lowest clearly perceptible current (sensory threshold) and the maximum to 0.5\,mA below visible contraction (motor threshold) and held constant within a trial \cite{corbet2018sensory}.


During the online protocol (Sessions 2--3), each session implemented real-time decoding of Start MI and Stop MI to control the exoskeleton. Upon the cue for Start MI, participants initiated MI; after successful decoding of Start MI, the exoskeleton began to reach toward a randomly selected target. Immediately after movement onset, the Stop MI decoder operated asynchronously (i.e., without an explicit stop cue). Participants were instructed to maintain the Start state during the reach and then intentionally transition to Stop MI when they judged their hand to be near the target, with the goal of halting the robot as close to the target as possible (Fig.~1, g), thereby testing whether they could volitionally titrate MI termination to achieve spatial precision. To balance task completion while minimizing user frustration, onset decisions were required within 5\,s of the green LED cue (otherwise declared a miss), and offset decisions within 6\,s of movement onset (otherwise declared a timeout), consistent with~\cite{tonin2021noninvasive}. If Stop MI was detected (\S II-C), the robot halted in place, then entered a rest phase and returned to home. If the Stop MI command was missed, the robot completed a pre-programmed 5\,s trajectory into the compliant (foam) target to allow benign over-travel, rested for 2\,s, and then returned.

To substitute for the continuous visual feedback commonly used in MI-BCI training, we mapped stNMES intensity to the BCI output: intensity was scaled between participant-specific thresholds proportional to the posterior probability (Eq. \ref{eq:EMA}). Triceps stimulation increased with Start MI evidence, whereas biceps stimulation increased with Stop MI evidence, reinforcing the intentional polarity via an agonist–antagonist pairing and obviating a screen \cite{biasiucci2018brain}.


\subsection{Harmony SHR\textsuperscript{\textregistered} Exoskeleton} \label{harmony}

We used the Harmony SHR\textsuperscript{\textregistered} exoskeleton (Harmonic Bionics, USA), a clinical-grade, bimanual upper-limb robot with seven degrees of freedom (DOFs) per arm—including five devoted to coordinated shoulder motion that preserve natural scapulohumeral rhythm. Harmony can assist, resist, or challenge upper-limb movements while logging end-effector pose, joint kinematics, and kinetic variables.

The BCI was integrated with Harmony over a low-latency UDP link so that the decoder directly gated robotic assistance in real time. The decoder output was a binary assistance gate, $g_{\text{bci}}\in\{0,1\}$: detection of Start MI set $g_{\text{bci}}=1$ to enable assistance, and detection of Stop MI set $g_{\text{bci}}=0$ to remove assistance—even mid-trajectory. When assistance was off, the controller continued to compensate for robot dynamics and enforce scapulohumeral coordination to keep the limb comfortable and stable. This created a simple neural-to-mechanical mapping for initiating and halting movement aid.

The joint-space control torque was composed as

\begin{equation}
    \tau_{\text{control}} = \tau_{\text{dyn}} + \tau_{\text{shr}} + g_{\text{bci}}\cdot \tau_{\text{task}}
    \label{eq: HarmonyTroque}
\end{equation}

\noindent
where $\tau_{\text{dyn}}$ compensates robot dynamics (e.g., gravity/weight support, Coriolis/centrifugal, and frictional effects), $\tau_{\text{shr}}$ enforces scapulohumeral coordination \cite{kim2017upper}, and $\tau_{\text{task}}$ is the assistive term that tracks a predefined reach trajectory with an impedance-based policy using conservative stiffness and damping to yield smooth motion and low apparent inertia. The BCI gate $g_{\text{bci}}$ modulated only the assistive channel. Thus, when the gate was on, the robot guided the arm along the desired path; when off, the motion ceased smoothly under intrinsic damping while posture remained stationary due to the coordination constraints.


\subsection{Training and Real-Time Decoding} \label{training}

We decode MI with a Riemannian-geometry pipeline and a minimum-distance-to-mean (MDM) classifier \cite{kumar2024transfer}. EEG was band-passed (8–30 Hz) and common-average referenced. From a sample of length 1 s (updated every 62.5 ms), we form a trace-normalized spatial covariance \cite{kumar2019towards}
\begin{equation}
    \Sigma_t \;=\; \frac{X_t^\top X_t}{\mathrm{trace}(X_t^\top X_t)} \in \mathbb{R}^{N_t\times N_c}
    \label{eq:trCov}
\end{equation}

\noindent
where $X_t\in\mathbb{R}^{N_t\times N_c}$ is the preprocessed data in the current buffer.
%
$\Sigma_t$ is a symmetric positive definite (SPD) matrix and thus lies on the SPD manifold. Using the affine-invariant Riemannian metric (AIRM) on this manifold, the distance between two SPD matrices can be estimated as  
\begin{equation}
\delta_R(\Sigma_1,\Sigma_2)\;=\;\big\|\log\!\big(\Sigma_1^{-\tfrac{1}{2}}\Sigma_2\Sigma_1^{-\tfrac{1}{2}}\big)\big\|_F
\label{eq:riemannDist}
\end{equation}
where $\log(\cdot)$ denotes the matrix logarithm and $\|\cdot\|_F$ the Frobenius norm. This distance is invariant under affine transformations and has been  used in
online settings \cite{kumar2024transfer}\cite{racz2024combining} for domain adaptation to mitigate EEG non-stationarity.

Using the above-mentioned distance, we can estimate the mean of a set of covariance matrices, also known as the Riemannian mean. For each class $k\in\{\text{Start},\text{Rest}\}$ or $k\in\{\text{Stop},\text{Maintain}\}$, we estimate its Riemannian mean ($\bar{\Sigma}_k $) from the corresponding training covariances $\{\Sigma_i^k\}$:
\begin{equation}
\bar{\Sigma}_k \;=\; \arg\min_{\Sigma\in\mathcal{S}_{++}}\;\sum_{i}\delta_R^2\!\big(\Sigma,\Sigma_i^k\big).
\end{equation}
Here, $\mathcal{S}_{++}$ denotes the set of SPD matrices. In practice, the Riemannian mean is computed using an iterative gradient descent–based method
\cite{zanini2017transfer}.


\textit{Session-wise recentering (distribution alignment).} \label{recenter}
To mitigate inter-session EEG drift, we recenter training and online covariances by their corresponding session means \cite{kumar2024transfer}. Let $S_{\mathrm{train}}$ be the Riemannian mean of all training covariances pooled across classes, and let $S_{\mathrm{online}}$ be a running Riemannian mean of recent covariances in the online session. Recentering is then performed using the square-root inverse of these means as a whitening transform on corresponding  training and online covariances. 



Distances for classification are then computed between the recentered class prototypes, $\hat{\Sigma}_{k,\mathrm{train}} = S_{\mathrm{train}}^{-\tfrac{1}{2}} \bar{\Sigma}_k S_{\mathrm{train}}^{-\tfrac{1}{2}}$, obtained from training covariances, and the transformed online covariance $ \hat{\Sigma}_{\mathrm{online}}= S_{\mathrm{online}}^{-\tfrac{1}{2}}\,\Sigma_{\mathrm{online}}\,S_{\mathrm{online}}^{-\tfrac{1}{2}}
$, estimated per sample using Eq. \ref{eq:trCov} from the online session \cite{kumar2024transfer}.

\textit{MDM score, posterior mapping, and smoothing.}
For a given decoder (Start vs Rest, or Stop vs Maintain), we compute AIRM distances $d_k(t)=\delta_R\!\big(\hat{\Sigma}_t,\hat{\Sigma}_k\big)$ using Eq. \ref{eq:riemannDist} and map them to pseudo-posteriors with a softmax over negative distances,

\begin{equation}
    p_k(t) \;=\; \frac{\exp\!\big(-\alpha\, d_k(t)\big)}{\sum_{j}\exp\!\big(-\alpha\, d_j(t)\big)}
    \label{eq:softMaxDist}
\end{equation}

\noindent
where the temperature $\alpha>0$ is chosen from the offline session. To suppress rapid fluctuations, posteriors are smoothed by an exponential moving average (EMA),
\begin{equation}
    \hat{p}_k(t) \;=\; (1-\beta)\,\hat{p}_k(t\!-\!1) \;+\; \beta\,p_k(t),
    \label{eq:EMA}
\end{equation}

\noindent
with $\beta\in(0,1)$ tuned offline.

\textit{Decision rule and thresholds.} \label{threshold}
Each decoder has a subject-specific decision threshold $\theta$ set from the offline session (ROC-based selection under a latency constraint).
A command is issued when the corresponding smoothed posterior first exceeds its threshold for  a short hold-time (0.25 s); a brief refractory period is enforced to prevent rapid toggling.

\textit{State machine and robot gating.} \label{state_robot}
Online, both decoders run in real time with a 1 s buffer updated every 62.5 ms. The Onset decoder is active from trial onset; when $\hat{p}_{\text{Start}}(t)\ge\theta_{\text{StartMI}}$ a Start decision is emitted and assistance is enabled, $g_{\text{bci}}(t) = 1.$ After movement onset, only the Stop decoder remains active; when $\hat{p}_{\text{Stop}}(t)\ge\theta_{\text{StopMI}}$ a Stop decision is emitted and assistance is halted, $ g_{\text{bci}}(t) = 0.$ If no threshold is reached within the decision window, the trial is a timeout; if the opposite class crosses first, the trial is a miss.


\subsection{Neurophysiological Analysis and Performance Metrics}

    \subsubsection{Time-Frequency Analysis} We
    computed time–frequency maps of EEG power throughout the trial  using offline data (Session 1).
    Data were segmented with a 0.5 s sliding window, advanced every $1/16$ s. Within each window, spectral power was estimated using Welch’s method and converted to a baseline-normalized index so that all frequencies are comparable. For each trial and frequency $f$, the baseline $B(f)$ was the average spectral amplitude during the 1 s resting period immediately preceding the countdown; the power was then normalized
    as
    
    \begin{equation}
        \mathrm{ERD/ERS}(f,t) = \log_{10}\!\big(A(f,t)/B(f)\big),
        \label{eq:ERD}
    \end{equation}

    \noindent
    where $A(f,t)$ is the windowed spectral amplitude. Negative values indicate ERD (power suppression) and positive values indicate event-related synchronization (ERS). Although ERD/ERS was computed at all channels, we present C3, the canonical EEG channel over left sensorimotor cortex contralateral to the right arm, and band-averaged ($8$–$30$ Hz) time courses. Subject-level maps were obtained by averaging the $\log_{10}$-ratios across trials; group-level maps were formed by averaging the resulting subject maps.


    \subsubsection{Online Performance}
    We analyzed the two online sessions (four runs per session, 20 trials per run). In every trial, Start MI was prompted. A hit was recorded if the Start posterior crossed its threshold within the decision window; a miss if the opposite class crossed first; and a timeout if no decision was reached. Stop MI was evaluated only on trials where Start had been successfully decoded (movement initiated), since termination can only be attempted once the robot is moving. For those
    trials, hit/miss/timeout were defined analogously using the Stop posterior and threshold. For each subject and session, we report the proportions of hit, miss, and timeout for onset and (attempted) offset; the primary online accuracy is the hit proportion.

    \subsubsection{Time-Delivery Accuracy} Onset decoding time was defined as the interval from the green LED cue (Start instruction) to the instant the Start posterior crossed its threshold. Offset decoding time was defined as the interval from movement onset (assistance enabled, $g_{\text{bci}}=1$) to the instant the Stop posterior crossed its threshold. Only hits contribute valid times.
    Timing is summarized per subject and session, and then at the group level to characterize central tendency and dispersion for onset and offset decisions.

\section{Results}

This section quantifies dual-state EEG–BCI performance using baseline-normalized time–frequency analysis of sensorimotor modulation, online onset/offset command-delivery outcomes (hit, miss, timeout), and decision latencies comparing cue-locked onset with movement-locked offset.

\subsection{EEG Modulations during the Task Paradigm}

\begin{figure}[t]
  \centering
  \includegraphics[width=\linewidth]{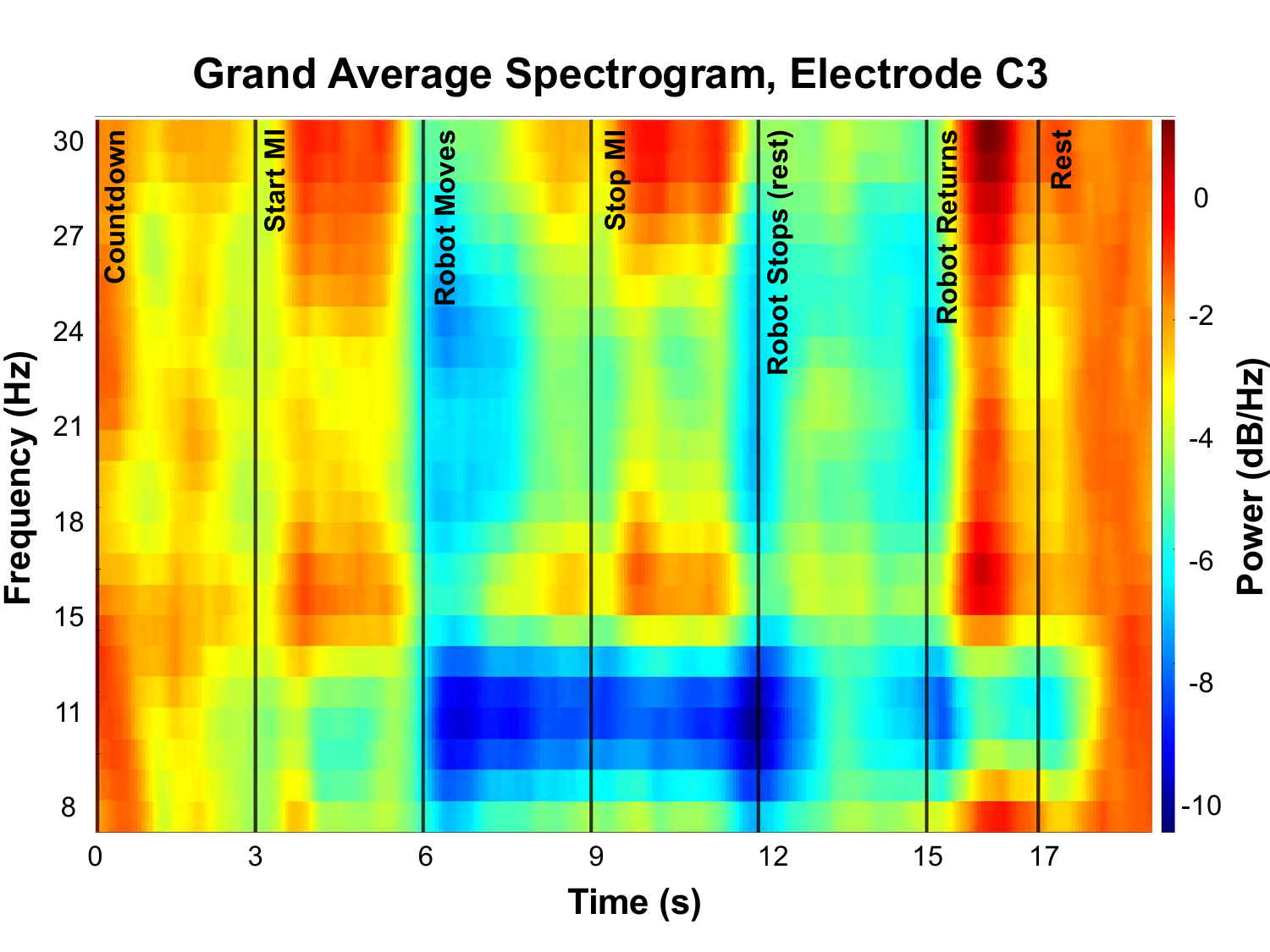}
  \caption{Grand average spectrogram from the C3 electrode showing changes in
spectral power over the course of task trials. Data were obtained by averaging
over all trials and all subjects. Vertical lines mark task events (countdown, Start MI, robot moves, Stop MI, robot stops, Rest, robot returns).}
  \label{fig:spectrogram}
  \vspace{-3mm}
\end{figure}

The grand-average spectrogram at channel C3 of offline data pooled across targets (Fig.~\ref{fig:spectrogram}) shows clear, event-related modulation of sensorimotor rhythms across the trial. During the countdown, spectral power remains near baseline. With the Start MI cue, a pronounced $\mu$-band ERD emerges and deepens throughout the Start MI period. As the robot begins to move, $\mu$ suppression persists with a modest depression extending into lower $\beta$, while a transient broadband elevation appears at higher frequencies—consistent with movement-related reafference. At the Stop MI cue, we observe a $\beta$-band ERS (increase in band power) that peaks shortly after imagery termination—the post-movement $\beta$ rebound—supporting that MI termination is a distinct process from passive rest and aligning with prior reports of a stop-specific signature \cite{orset2021stopping}. After the robot stops and rest ensues, power trends back toward baseline. During the return-to-home phase, elevated high-frequency activity co-occurs with a lingering $\mu$ ERD, reflecting proprioceptive/movement influences during the return trajectory. Similar ERD/ERS patterns were observed for all three target directions.

\subsection{Online Command Delivery Accuracy}

\begin{figure}[t]
  \centering
  \includegraphics[width=\linewidth]{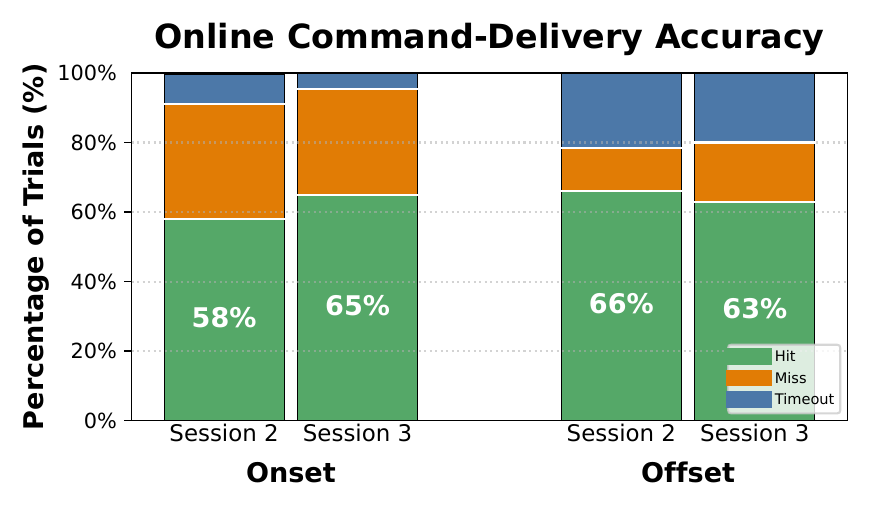}
  \caption{Online command-delivery outcomes. Stacked bars show group-mean outcome proportions (hit, miss, timeout) for onset and offset in Sessions 2–3; numbers indicate mean hit rate. Offset is computed on attempted trials only (i.e., trials following a successful onset).}
  \label{fig:cmd-del}
  \vspace{-3mm}
\end{figure}

Group performance shows reliable dual-state control with distinct error profiles for onset and offset (Fig. \ref{fig:cmd-del}). For Start MI (onset), the mean hit rate improved from 58\% in Session 2 to 65\% in Session 3. Errors were mostly misses (Session 2: 33.1\%; Session 3: 30.5\%), with timeouts rare (Session 2: 8.4\%; Session 3: 5.9\%). This
indicates that participants generally produced a decisive class output when prompted, with fewer trials stalling without a decision in Session 3.

For Stop MI (offset), computed only on trials where a Start hit initiated movement, the mean hit rate was 66\% in Session 2 and 63\% in Session 3. Here, errors skewed toward timeouts rather than misses (Session 2: timeouts 22.0\%, misses 12.4\%; Session 3: timeouts 20.2\%, misses 17.0\%), reflecting the inherently more demanding asynchronous stopping task. Despite being issued during movement, offset performance remained comparable to onset, demonstrating consistent real-time initiation and termination using noninvasive EEG.

\subsection{Timing Dynamics}


Fig.~\ref{fig:decode-time} summarizes decision-latency distributions for Start MI (onset; cue-locked) and Stop MI (offset; movement-locked to the onset of assistance), defined as the time from cue/movement onset to decoder threshold crossing (hits only). Onset decisions clustered around $\sim$1\,s, with mean $\pm$ SD of $0.990 \pm 0.953$\,s in Session~2 and $1.061 \pm 1.001$\,s in Session~3. Offset decisions were predictably slower---as designed---because participants were instructed to stop nearer the target: $3.419 \pm 1.050$\,s in Session~2 and $3.328 \pm 0.998$\,s in Session~3. These latencies closely match the theoretical optimum ($\sim$3.2\,s) implied by the reaching trajectories in the offline spectrogram timeline (Fig.~\ref{fig:spectrogram}).
Most onset commands occurred within $\sim$1--1.5\,s, whereas offset commands concentrated around $\sim$3--4\,s with occasional long-latency trials. Session-to-session changes were modest relative to within-session variability.

\section{Discussion}

This study shows that a noninvasive BCI can start and stop assistance during ongoing movement of a multi-DOF upper-limb exoskeleton in real time. Framed this way, the interface is a simple intent-gated ``go/stop'' control well aligned with rehabilitation needs: an intentional start to launch a functional reach and an intentional stop to terminate near a goal, offering a practical middle ground between continuous control and preprogrammed motion.

The neurophysiology supports this mapping. The spectrogram (Fig.~\ref{fig:spectrogram}) shows robust $\mu$-band desynchronization with Start MI and a clear $\beta$-band rebound after Stop MI. Crucially, these canonical signatures persist despite movement-related sensory feedback and instrumental noise: participants still produced $\mu$-ERD at onset and $\beta$-ERS at offset. The rebound is somewhat attenuated---consistent with the arm being moved and additional sensor noise partially masking the response---but the separation of states remains evident. That separation justifies dedicated onset and offset decoders rather than thresholding a single stream.

\begin{figure}[t]
  \centering
  \includegraphics[width=\linewidth]{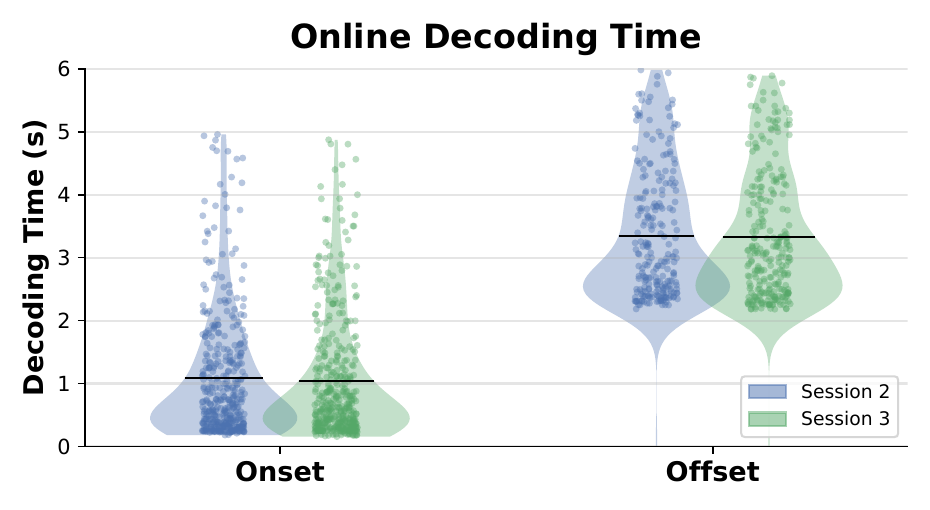}
  \caption{Online decoding time for hits only. Violin plots show the distribution of decision latencies for onset (cue-locked) and offset (movement-locked to assistance onset) in Sessions 2-3; black bars indicate means.}
  \label{fig:decode-time}
  \vspace{-3mm}
\end{figure}

Online results show that onset command delivery improved between sessions and showed few timeouts (Fig.~\ref{fig:cmd-del}), while offset achieved comparable hit rates but with errors skewed toward timeouts rather than misses. This pattern fits the instructions---participants attempted offset near the target---placing decisions close to the boundary; they typically either hit or timed out, with comparatively fewer outright misses. Decision-time distributions (Fig.~\ref{fig:decode-time}) are consistent: onset latencies cluster near $\sim$1~s, whereas offset is significantly slower as users time termination to the target rather than halting reflexively.

A notable limitation of the online results warrants discussion.
Online BCI performance is acceptable but far from optimal. 
This is likely caused by the recentering process that transforms the online EEG samples to mitigate EEG drift over sessions.
Next, we examine the effects of the conventional recentering process that, due to the nature of the experimental protocol, creates a bias in the distribution of transformed online EEG samples with respect to the class prototypes. We then introduce a new recentering method.


\subsection{Recentering Limitations and Pseudo-Online Alternatives} \label{pseudo}

Our recentering approach estimates a whitening reference $S$ from the online EEG samples processed by the BCI---hence, task-based.
This reference is then used to transform incoming covariances, and Riemannian distances are computed between the transformed covariance and the class prototypes. Operationally, this method seems natural: it attempts to align the distribution of covariances in the online session with that of the offline training data to reduce session-related shifts. However, unlike conventional BCI scenarios—where the reference is derived from the overall distribution of both positive (Start MI, Stop MI) and negative (Rest, Maintain-MI) classes—our online protocol only provides samples from the positive (target) class. As a result, the recentering operation effectively pulls the positive-class distribution toward the global center of the offline data, which includes both classes. This induces a bias that shifts positive samples closer to the offline mean, thereby increasing the likelihood of misclassification as the negative class.


\begin{figure}[t]
  \centering
  \includegraphics[width=\linewidth]{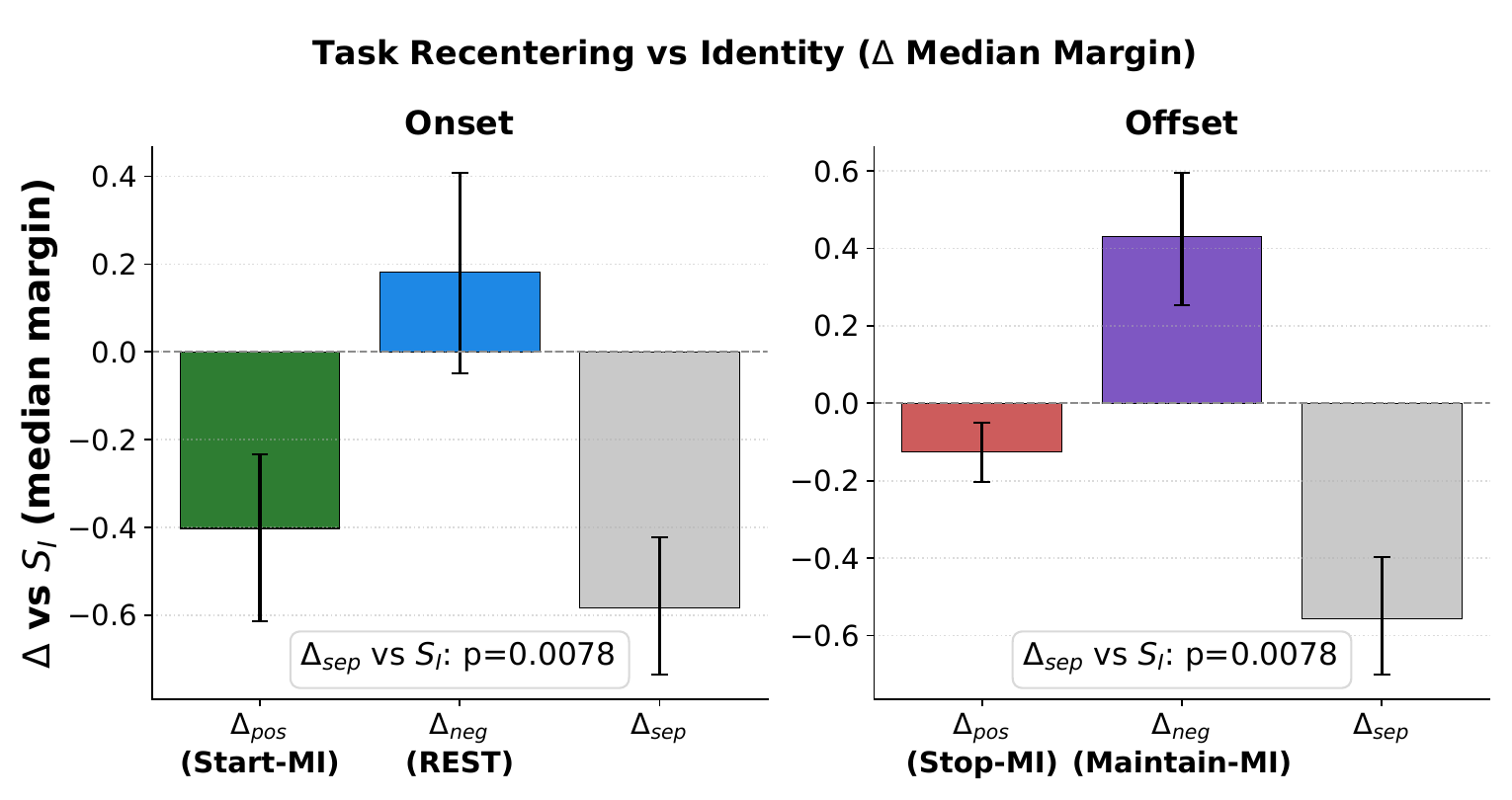}
  \caption{Task-based recentering shifts class separation relative to $S_i$ ($\Delta$ median margin). Bars show the group-mean change in median margin, $m=d_{negative}-d_{positive}$, under task-based recentering versus the identity reference $S_i$. Error bars denote $95\%$ CIs across subjects (subject means over runs).}
  \label{fig:bias}
  \vspace{-3mm}
\end{figure}

To diagnose and visualize how task-based recentering biases the data, we quantify the role of each sample with a distance-difference score that tells us which class the recentered sample most resembles. We first build two class prototypes from offline data: a positive prototype (Start MI for the onset decoder; Stop MI for the offset decoder) and a negative prototype (Rest for onset; Maintain for offset). During online decoding, for any incoming covariance $\Sigma_i$, we measure two Riemannian distances (Eq. \ref{eq:riemannDist}): $d_{\text{positive}}$ (from $\Sigma_i$ to the positive prototype) and $d_{\text{negative}}$ (to the negative prototype). We then compute a single margin,
\begin{equation}
    m \;=\; d_{\text{negative}} \;-\; d_{\text{positive}},
    \label{eq:margin}
\end{equation}

\noindent
so $m>0$ means the sample is closer to the positive class, and $m<0$ means it is closer to the negative class.

To analyze the effects of task-based recentering,
we also compute the margins using the identity reference (no whitening; geometry unchanged, $S_i$), referred to as identity-based recentering. Then, we examine how the median margin shifts for the two references. We define the margin shift for the positive samples and negative samples separately as, 
\begin{equation}
    \Delta_{\text{pos}}
    \;=\;
    \bar{m}_{\text{pos}}(\text{task})
    \;-\;
    \bar{m}_{\text{pos}}(\text{identity})
    \label{eq:pos_margin}
\end{equation}
\begin{equation}
    \Delta_{\text{neg}}
    \;=\;
    \bar{m}_{\text{neg}}(\text{task})
    \;-\;
    \bar{m}_{\text{neg}}(\text{identity})
    \label{eq:neg_margin}
\end{equation}

\noindent
We define a single metric capturing the change between classes,
\begin{equation}
    \Delta_{\text{sep}}
    \;=\;
    \Delta_{\text{pos}} \;-\; \Delta_{\text{neg}}
    \label{eq:delta_margin}
\end{equation}

\noindent
Fig.~\ref{fig:bias} visualizes these values. For both onset and offset, $\Delta_{\text{pos}}<0$ and $\Delta_{\text{neg}}>0$, so $\Delta_{\text{sep}}$ is strongly negative (paired Wilcoxon $p=0.0078$). This means, when the reference is built from Start MI, Start MI samples’ margins drop (they look less distinctly Start relative to Rest), while Rest samples’ margins rise (they look more Start-like). When the reference is built from Stop MI, Stop MI samples’ margins drop, while Maintain samples’ margins rise. In both cases, the positive and negative score distributions move closer together, so they are less separable. With a single fixed threshold, more samples fall in the ambiguous middle, making decisions less reliable and degrading online BCI performance over time.

However, if we remove recentering altogether, we fail to capture day-to-day EEG drift.
Therefore, we propose a new recentering technique that uses a fixation-based reference built only from the pre-cue fixation period before each trial (Fig.~\ref{fig:setup}, e).
Critically, this reference uses samples that are class agnostic---i.e., these EEG samples
belong neither to the positive nor the negative class.

The method works as follows. Using online fixation samples, the method performs five operations (i) compute covariances (Eq. \ref{eq:trCov}) and a log-Euclidean mean; (ii) trim outliers on the SPD manifold—intuitively, we reject fixation samples that contain blinks or movement artifacts by removing the farthest $~20\%$ from the initial mean, measured by Eq. \ref{eq:riemannDist}, and then recompute the mean; (iii) gently shrink the result toward the identity in the log domain, which regularizes the reference toward a neutral, class-agnostic distribution,
\begin{equation}
    \log S_{\mathrm{fix}} = (1-\alpha)\,\log S_{\mathrm{raw}}
    \label{eq:log_fix}
\end{equation}

\noindent
where $S_{\mathrm{raw}}$ is the unregularized fixation mean, $\alpha$ is the identity-shrink strength (per run); (iv) apply eigenvalue (diagonal) shrinkage for conditioning, so no single channel dominates and distances remain stable,
\begin{equation}
    S_{\mathrm{fix}} = V \,\mathrm{diag}\!\big( (1-\lambda)\,\boldsymbol{\ell} + \lambda\,\bar{\ell}\,\mathbf{1} \big)\, V^{\top}
    \label{eq:fix_diag}
\end{equation}

\noindent
where $V$ and $\ell$ are the eigenvectors and eigenvalues of the current $S_{\mathrm{fix}}$, $\bar{\ell}$ is the mean eigenvalue, and $\mathbf{1}$ is a vector of ones; and (v) smooth across runs in the log domain to slowly track the drift,
\begin{equation}
    \log S_{\mathrm{fix}}^{(r)} = (1-\beta)\,\log S_{\mathrm{fix}}^{(r-1)} + \beta\,\log S_{\mathrm{raw}}^{(r)}.
    \label{eq:fix_smooth}
\end{equation}

\noindent
where $\beta$ is the across-run smoothing weight, with $r$ indexing the run.

We used grid search to select hyperparameters: $\text{outlier rejection percentage}=0.20$, $\alpha=0.25$, $\lambda=0.05$, $\beta=0.30$ (requiring $N_{\min}=8$ fixation samples to form a new reference; otherwise we reuse the prior run’s reference). Together, these choices suppress outliers, keep the transformed online EEG samples in a neutral (class-agnostic) distribution, and adapt smoothly within and across sessions—precisely the behavior missing from task-based recentering.

\begin{figure}[t]
  \centering
  \includegraphics[width=\linewidth]{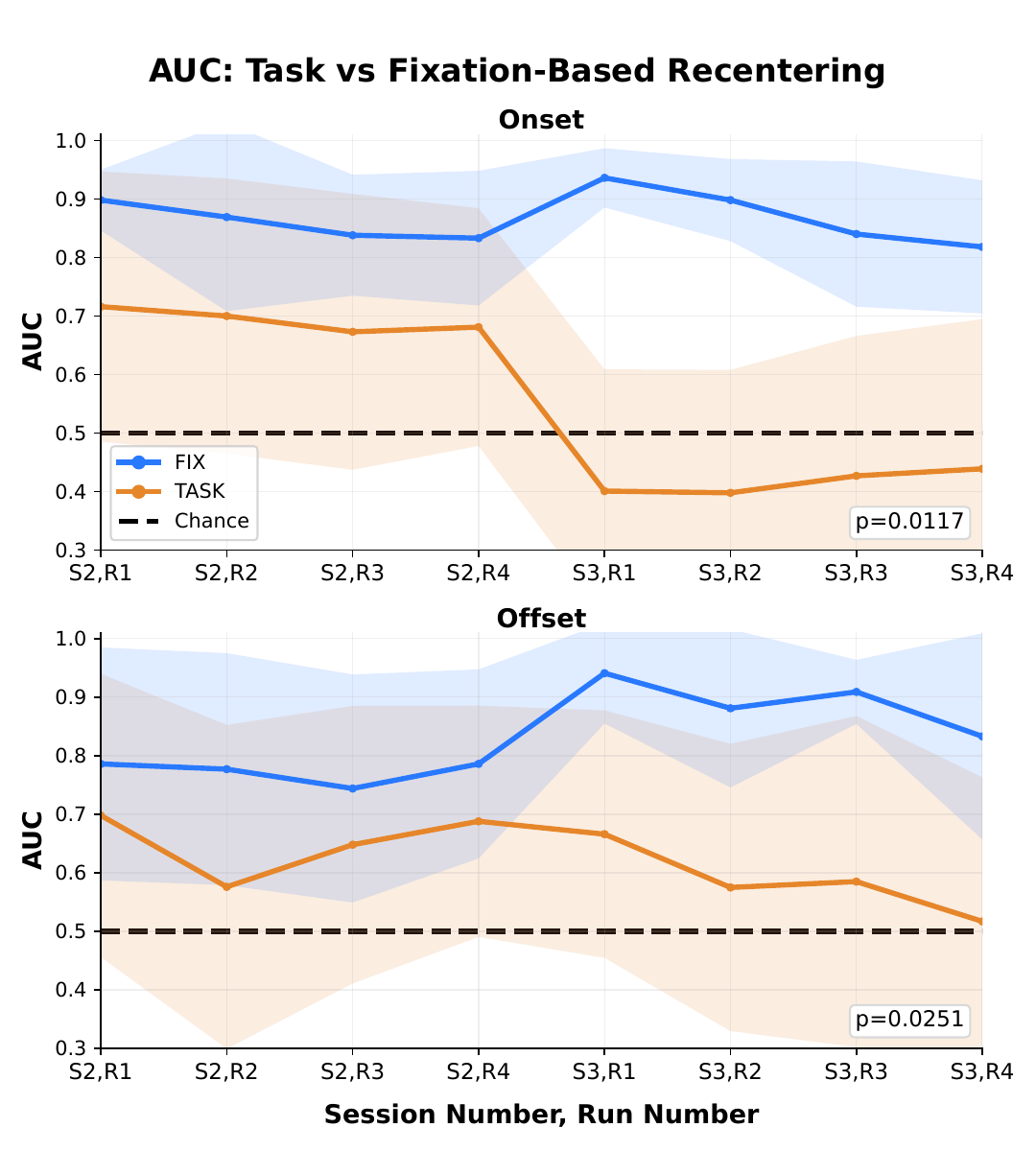}
  \caption{AUC by run for onset and offset under task- vs fixation-based recentering. Lines show group-mean AUC; ribbons denote 95\% CIs. Inset boxes report paired Wilcoxon $p$-values for per-subject mean AUC differences (Task$-$Fix).}
  \label{fig:AUC}
  \vspace{-3mm}
\end{figure}

To test the new $S_{\mathrm{fix}}$, we ran a pseudo-online analysis on the recorded online data. Each sample was recentered in two ways--once with the task-based reference and once with the fixation-based reference—and we recomputed the distance-difference margin $m$. We then computed AUC per run for each recentering approach; because AUC is threshold-free, it is a fair comparison under EEG drift.

Fig.~\ref{fig:AUC} shows the results. With the fixation-based reference, onset AUC rises from $0.554\pm0.291$ (task) to $0.866\pm0.073$ (fix), a $+56\%$ relative gain (paired Wilcoxon $p=0.0117$). Offset AUC rises from $0.619\pm0.266$ to $0.832\pm0.124$, a $+34\%$ gain ($p=0.0251$). Many per-run comparisons are also significant. Critically, across the Session-2 to Session-3 boundary (S2-R4 to S3-R1), fixation-based AUC stays high while task-based AUC drops, indicating markedly better tolerance to between-day EEG drift.

We also recomputed the same distance-difference summary from above as a single “separation change” $\Delta_{\text{sep}}$. Fixation reduces the one-sided shift substantially. At onset: task $=-0.584\pm0.242$ vs.\ fix $=-0.442\pm0.206$ (improvement $\Delta(\text{Fix}-\text{task})=+0.142\pm0.079$, a $24.4\%$ reduction; $p=0.0117$, $n=8$). At offset: task $=-0.556\pm0.239$ vs.\ fix $=-0.451\pm0.199$ (improvement $+0.105\pm0.076$, an $18.9\%$ reduction; $p=0.0117$). Taken together, fixation-based recentering both increases separability and suppresses unilateral bias, yielding a more symmetric, more stable decision boundary that is robust within runs and across days.

\subsection{Limitations and Future Work}

Despite encouraging results, this study remains an initial feasibility demonstration with important limitations. The cohort was small, all participants were healthy, and each completed only two online sessions, limiting our ability to assess learning, longer-term adaptation, and the development of reliable offset control. Thus, the present findings show that online, intention-gated start/stop control is achievable under controlled conditions, but do not yet establish readiness for rehabilitation use. As an immediate next step, we will evaluate the fixation-based recentering method in a fully online, multi-session setting to test whether repeated exposure improves stopping precision, consistency across days, and overall control stability.

A key translational limitation is that the system has not yet been evaluated in the target rehabilitation population. Post-stroke users will likely introduce greater heterogeneity, including altered sensorimotor rhythms, increased non-stationarity, and fatigue, all of which may affect calibration and online reliability. While the fixation-based recentering may help mitigate some of this variability, this must be established in patient studies. Future work will therefore extend evaluation to post-stroke participants and incorporate measures such as stop-accuracy kinematics, reduced system latency, and improved characterization of false activations. Together, these steps will determine whether the present framework can progress from healthy-subject feasibility toward clinical translation.

\section{Conclusions}

We have demonstrated online, intention-gated start/stop control of a clinical upper-limb exoskeleton with a noninvasive BCI, bridging MI transition decoding and real-time robot assistance. A geometry-aware Riemannian pipeline with subject-specific thresholds and proprioceptive feedback enabled reliable multi-session operation: participants consistently initiated assistance and terminated it mid-trajectory in real time. Methodologically, we revealed a bias from task-based recentering and quantified its asymmetric margin shift. Replacing it with a class-agnostic fixation reference that tracks EEG drift while preserving class geometry yielded substantially higher threshold-free separability and reduced unilateral bias within and across days. Together, these advances establish an important feasibility step toward MI-BCI rehabilitation by enabling precise initiation and termination with reduced recalibration burden, while motivating further evaluation in patient populations before clinical translation.



\bibliographystyle{ieeetr}
\bibliography{ref}

\end{document}